\documentclass[11pt]{article}
\usepackage{dialogue2021}
\usepackage{booktabs}
\usepackage{ifxetex}

  \usepackage[T2A]{fontenc}
  \usepackage[utf8]{inputenc}
  \usepackage{cmap}
  \usepackage{times}
  \usepackage{latexsym}

\usepackage[british,russian]{babel}
\usepackage{url}
\usepackage{pgf}

\usepackage{covington} 

\newcommand\citep[1]{\cite{#1}}

\dialogfinalcopy 

\title{SemSketches-2021: experimenting with the machine processing of the pilot semantic sketches corpus}

\author{
Maria Ponomareva$^{\spadesuit, \heartsuit}$~~~
Maria Petrova$^{\spadesuit}$~~~
Julia Detkova$^{\spadesuit}$~~~ 
Oleg Serikov$^{\heartsuit,  \diamondsuit }$~~~
Maria Yarova$^{\clubsuit}$~~~\\
$^{\spadesuit}$ABBYY, Moscow, Russia \\[1mm]
$^{\heartsuit}$National Research University Higher School of Economics, Moscow, Russia\\[1mm]
$^{\clubsuit}$Moscow Institute of Physics and Technology, Moscow, Russia\\[1mm]
$^{\diamondsuit}$Deeppavlov MIPT, Moscow, Russia\\
}
 
\begin{document}
\selectlanguage{british}
\maketitle
\begin{abstract}
The paper deals with elaborating different approaches to the machine processing of semantic sketches. It presents the pilot open corpus of semantic sketches. Different aspects of creating the sketches are discussed, as well as the tasks that the sketches can help to solve. Special attention is paid to the creation of the machine processing tools for the corpus. For this purpose, the SemSketches-2021 Shared Task was organized. The participants were given the anonymous sketches and a set of contexts containing the necessary predicates. During the Task, one had to assign the proper contexts to the corresponding sketches.


\end{abstract}
 
\selectlanguage{russian}
\begin{center}
  \russiantitle{SemSketches-2021: опыт автоматической обработки пилотного корпуса семантических скетчей}

  \medskip \setlength\tabcolsep{0.1cm}
  \begin{tabular}{lllll}
    \textbf{Мария Пономарева} & \textbf{Мария Петрова}& \textbf{Юлия Деткова}& \textbf{Олег Сериков}& \textbf{Мария Ярова}\\
      ABBYY, ВШЭ & ABBYY& ABBYY&ВШЭ, Deeppavlov& МФТИ\\
      Москва & Москва& Москва & Москва & Москва  
  \end{tabular}
  \medskip
\end{center}

\begin{abstract}

Статья посвящена различным подходам к автоматической обработке семантических скетчей. В статье представлен первый открытый корпус семантических скетчей для русского языка. На примере данного корпуса рассматриваются особенности семантических скетчей и проблемы, возникающие при их построении, обсуждаются задачи, которые могут решаться с привлечением скетчей, а также дальнейшие перспективы использования скетчей.
Особое внимание уделяется возможности создания инструментов автоматической обработки корпуса. В качестве эксперимента по созданию подобных инструментов авторами проведено соревнование  SemSketches-2021, в рамках которого участникам предлагалась задача по работе с корпусом скетчей, где требовалось соотнести анонимизированные скетчи с рядом контекстов для соответствующих предикатов. 

\medskip

\textbf{Ключевые слова:} скетчи слов, семантические скетчи, семантика фреймов, разметка семантических ролей, корпусная лексикография
\end{abstract}

\selectlanguage{english}
\section{Introduction}
The current paper continues the work on the semantic sketches which were first presented at the Dialogue-2020 conference. 

The idea of the semantic sketch was introduced in \citep{Detkova2020}. The semantic sketch is a special representation of a word’s compatibility where all semantic links of the word are grouped according to their semantic relations with the core they depend on. All possible semantic dependencies are statistically ranged: first, the frequency of the collocation between the parent and the child is taken into account; second, the frequency of the semantic role for the given core (for instance, the frequency of the Agent, Locative, Object, or Time). 

The most frequent collocations form the semantic sketch of the word.
In \citep{Detkova2020}, the authors focused on the creation of the semantic sketches and on testing the semantic mark-up used for the sketches. Namely, they measured the correctness of the predicate’s choice in a set of sentences and the choice of the proper semantic roles for the predicates’
 dependencies.

In the present work, the focus has been made on building the pilot corpus of the semantic sketches themselves, \textbf{the SemSketches corpus}. The corpus is aimed at achieving several purposes: 

\begin{enumerate}
\item to evaluate how representative the sketches are,
\item to elaborate some tools for processing the sketches, 
\item to specify what kind of tasks the semantic sketches can help to solve, as our further plan is to integrate the sketches into the General Internet-Corpus of Russian (GICR, \citep{belikov13},  \citep{piperski13}),
\item  to analyze what kind of mistakes we happen to face while creating the sketches.

\end{enumerate}
The idea to represent a word’s meaning in the form of the semantic sketch is closely related to the main idea of distributional semantics according to which the meaning of the word can be represented through its lexical co-occurrence. The famous formula for the idea  given in \citep{Firth1957} says: ``You shall know a word by the company it keeps''.

Over the past few years, vector representations have become a traditional method of representing the word's semantics. The static embeddings such as word2vec~\citep{Mikolov13} and FastText ~\citep{bojanowski2017enriching} as well as the dynamic embeddings that followed, such as  ELMo ~\citep{peters-etal-2018-deep}, ULMFit ~\citep{Howard}, and BERT ~\citep{devlin-etal-2019-bert}, have completely changed the NLP field. However, quality evaluations of the vector representations pose a challenge, as their serious drawback is that one can neither  assess nor interpret them directly. 

Whereas the vector is a numeric meaning representation, appropriate for computers, the semantic sketch can be considered  its human-interpretable counterpart. 

As an experiment on processing the sketches automatically, we have introduced \textbf{the SemSketches Shared Task}. One of its goals is to connect these two methods of semantic representation.

The Shared Task suggested the following problem.
Participants were given the corpus of the semantic sketches with the core predicates unknown, that is, the semantic roles of the dependencies and the word-fillers of the roles were given, but not the predicates they were attached to. We have presented a set of such anonymous sketches and a list of contexts containing the predicates. The task was to create a tool that assigns the sketch to the corresponding contexts. 

For most sketches, the task did not seem difficult for a human, as some of the examples will demonstrate below, but it turned out to be rather complicated for the computer, as the results of the competition showed.
The corpus and the Shared Task results are available at the SemSketches github\footnote{\url{https://github.com/dialogue-evaluation/SemSketches}}.

\section{What is a semantic sketch}
There is no need to underline the importance of using text corpora for various purposes nowadays. The size of the corpora is growing quickly. On the one hand, it gives the users more opportunities and allows one to receive more representative data. On the other hand, with a bigger corpus, more sophisticated tools are demanded to process the results of the search queries.

One of the methods to describe the word’s compatibility is to present it in the form of a syntactic sketch \citep{kilgarriff2014sketch}. The syntactic sketch is a lexicographical profile of a word, where word dependencies are classified by their grammatical roles and ranged by the statistics of their compatibility with the core. The syntactic sketches were first introduced in the Sketch Engine project\footnote{\url{www.sketchengine.eu}} and over the past years have become widely used in lexicography, language teaching, multilingual corpora creation, various translation resources, and in a number of other areas. 

The evident advantage of the syntactic sketch is its vividness: it shows simultaneously all of the most frequent syntactic dependencies of a word and arranges them in a table according to the roles. 
At the same time, the syntactic sketches have one strong limitation: the grammatical information they are based on does not allow one to take lexical homonymy into account, which complicates the interpretation of the obtained results.

In order to solve this problem, an attempt was made to create the semantic sketches \citep{Detkova2020}, where the representation of a word’s compatibility is supplemented with semantic relations between words (each relation is marked not only with a syntactic, but with a semantic role as well) and semantic classes of words (which mark the specific semantic meaning of a word in a context).

Therefore, the semantic sketch is understood as a generalized lexicographic portrait of a word, which includes the most frequent semantic dependencies of the verb. In other words, it is a way of representing the compatibility of words, where the description of each word includes a set of its most frequent semantic dependencies classified according to their semantic roles. For each role a number of relevant ``fillers'' (words and phrases) are given, and the fillers are ranked according to the frequency of their compatibility with the core. Each sketch illustrates a word with a certain meaning.

The semantic sketches are built with the help of the Compreno parser \citep{Compreno2012}. Unlike other parsers, the Compreno suggests full semantic mark-up, namely, it deals not only with the actant semantic dependencies of the predicates, but with the adjuncts, modifiers, and other dependencies as well \citep{Petrova2013}.
It makes the sketches an important tool for dealing with the semantic role labeling (SRL) problem which has attracted many researchers recently.  

Despite high interest in the problem (\citep{gildea02}, \citep{koomen-etal-2005-generalized}, \citep{PalmerMarthaStone2010Srl}, \citep{lang-lapata-2011-unsupervised-semantic},
\citep{DBLP:conf/aaai/TanWXCS18},
\citep{cheng-etal-2017-learning},
\citep{he-etal-2018-syntax}), until the current moment no research among the SRL papers has been presented (or, at least, we have not seen any), where all semantic roles are taken into account. Most works focus on the actant dependencies only, such as Agent, Object, or Experiencer.
In the meantime, for many predicates, circumstantial dependencies are enough frequent and significant to get into the predicate’s sketch together with its actants, and, moreover, in some cases, help to identify the core even better than the actants do.
The sketches are illustrated in the two examples below, the first  one --- for the verb 
\selectlanguage{russian}
<<страдать  :SUFFERING\_AND\_TORMENT>>
\selectlanguage{british}
‘to suffer’ (Figure \ref{fig:suffer}) and the second one --- for the verb 
\selectlanguage{russian}
<<готовить:TO\_PREPARE\_FOOD\_SUBSTANCE>>
\selectlanguage{british}

‘to prepare food, to cook’

\begin{figure*}[ht]
  \centering
  \includegraphics[width=\textwidth]{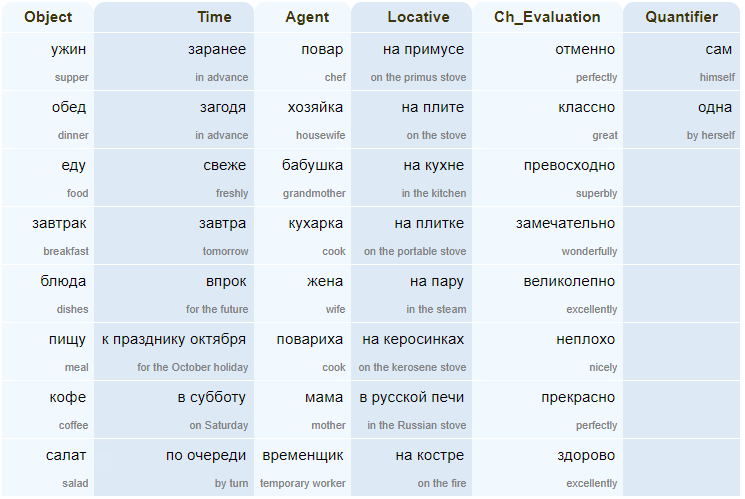}
  \caption{the sketch for the verb 
  \selectlanguage{russian}
  <<готовить:TO\_PREPARE\_FOOD\_SUBSTANCE>>
\selectlanguage{british}
(‘to prepare food, to cook’). Here the elements of the sketch are given with their rough translations.}
  \label{fig:prepare}
\end{figure*}

(Figure \ref{fig:prepare}):

\begin{figure*}[ht]
  \centering
  \includegraphics[width=\textwidth]{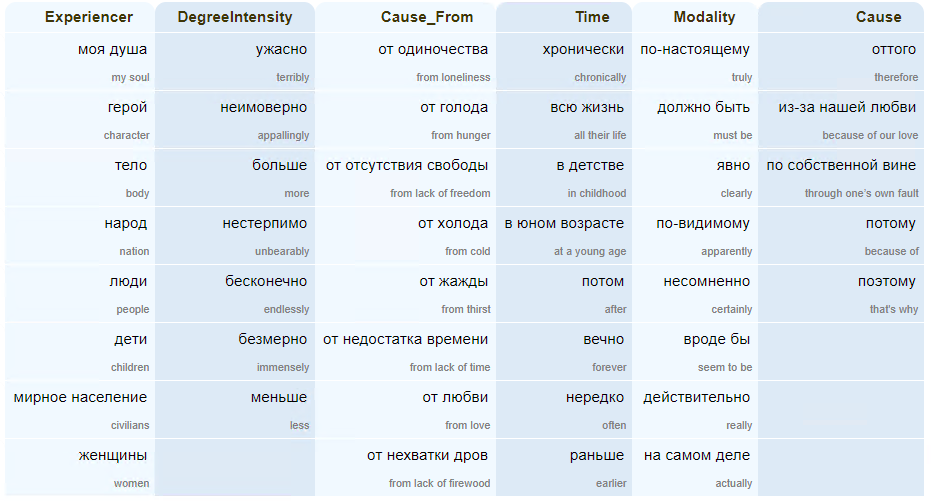}
  \caption{the sketch for the verb 
  \selectlanguage{russian}
  <<страдать:SUFFERING\_AND\_TORMENT>>
  \selectlanguage{british}
(‘to suffer’). Here the elements of the sketch are given with their rough translations.}
  \label{fig:suffer}
\end{figure*}

The participants of the Shared Task got the same representations, but did not  get the titles of the sketches. However, as the pictures above demonstrate, it does not seem difficult for a human to guess the proper predicates for the sketches, which allows us to regard the sketches as representative  illustrations for the verb’s compatibility.

\section{The SemSketches Shared Task}
To explore the semantic sketches as far as their quality and representativeness are concerned, we have created the pilot corpus of Russian semantic sketches and made it the basis for the SemSketches Shared Task. 
The problem was formulated as follows: \textit{given a set of anonymized sketches and a set of contexts for different predicates, one should match each predicate in its context to a relevant sketch.}

The second part of the competition data is the set of the contexts given for different predicates. In the case of ambiguous predicates, the WSD problem can be stated.

\subsection{Data preparation}

\subsubsection*{Sketches}
The sketches were built on the texts from the Magazine Hall of the GICR.

Although the parser gives us the full semantic mark-up, we have implemented some restrictions for the present research. As in \citep{Detkova2020}, the authors have taken only verbal cores and their subtrees: all verbs are marked with semantic classes (denoting their meanings) and the semantic roles for their direct dependencies.

We did not mark the dependencies of the non-vebal cores, the dependencies of the ellipted verbs and the ellipted groups themselves, as well as the syntactically moved groups.
In addition, we have introduced some additional restrictions for the purpose of the current competition, namely, we have excluded pronouns and personal nouns, as they complicate the work with the anonymized sketches.

For the current corpus, we have chosen only verbs which have at least two meanings, as it makes the task of defining proper sketches more interesting, on the one hand, and, on the other hand, contributes to solving the WSD problem. It means that each verb chosen entered at least two semantic classes.

The number of such verbs for the Russian language turned out to be more than ten thousand. Then we chose a subset of the list through selecting the verbs by the following principles. 

At the beginning, we have ranged the sample so that the verbs with the most frequent meanings came first: for instance, the verb 
\selectlanguage{russian}
\textit{рубить} meaning TO\_HACK (\textit{\textbf{рубить} дерево} 
\selectlanguage{british}
--- `to hack a tree') is sufficiently frequent, while the same verb meaning TO\_KNOW\_ABOUT 
\selectlanguage{russian}
(\textit{\textbf{рубить} в математике}
\selectlanguage{british}
--- `to understand mathematics well') is rather marginal and has thus been positioned at the end of our list. The frequency of different meanings has been obtained with the help of the Compreno parser. 

Next, we have collected the verbs’ sketches taking into account the number of the relations the verb has in the corpus. 
Namely, we have collected all the semantic dependencies for each meaning of each verb in our marked-up corpus, and if the number of the dependent nodes exceeded the threshold of 2000, the predicate in the certain value was selected for inclusion in the final set. 
During this procedure, all dependencies were taken into account --- both different and repeated, in order not to lose any frequent predicates with limited lexical compatibility. At the same time, the threshold was rather high to preserve the quality of the sketches.

At last, the final number of sketches in the pilot corpus became 915.
Due to the exclusion of rare meanings, some verbs kept only one meaning in the sample, that is, the terminal verb list contained both polysemantic verbs with several meanings in the sample and polysemantic verbs which entered in our sample only in one (the most frequent) meaning.

The next step was to analyze the correctness of the sketches, namely, to check whether the semantic dependencies and the fillers of the dependencies that got into the sketch really refer to the verb in the given meaning. The errors check was performed for a subsample of the corpus which formed the manual Dev data (see below).

Most errors refer to situations where the more frequent homonym influences the less frequent one. For instance, the verb 
\selectlanguage{russian}
\textit{писать}
\selectlanguage{british}
meaning `to paint' (
\selectlanguage{russian}
\textit{\textbf{писать} портрет с кого-л.} 
\selectlanguage{british}
---
`to paint smb.'s picture'
)
is less frequent than 
\selectlanguage{russian}
\textit{писать} meaning `to write' (
\textit{\textbf{писать} письмо}
\selectlanguage{british}
--- `to write a letter'), so the sketch for the
\selectlanguage{russian}
\textit{писать}
\selectlanguage{british}
--- `to paint' contains some incorrect examples in the Object dependency --- such as `to write letters'. 

The reason is that when building the semantic structures for the sentences the sketch is based on, the structure with the incorrect but more frequent homonym gets a higher evaluation due to the high statistics of the more frequent verb.

Another error can be illustrated with the sketch
\selectlanguage{russian}
<<готовить:TO\_PREPARE\_MEDICINE\linebreak[4]{}\_OR\_FOOD>> 
\selectlanguage{british}
`to cook'. It contains combinations like 
\selectlanguage{russian}
\textit{\textbf{готовить} резервную копию}
\selectlanguage{british}
--- `to prepare a reserve copy'. Here the problem is that the compatibility of `copy' with the verbs depends not on the `copy' itself but on the semantics of the noun following it, that is, `the copy of the cake' is also possible.

As an instance of the sketch with the incorrect semantic dependency, let us take the sketch 
\selectlanguage{russian}
<<выходить:идти:TO\_WALK>>
\selectlanguage{british}
`go out' on the Figure \ref{fig:walk}. The sketch contains the Agent\_Metaphoric slot which must be definitely referred to another meaning, and the Purpose\_Goal slot contains the incorrect filler
\selectlanguage{russian}
\textit{на связь} (\textit{выйти \textbf{на связь}}
\selectlanguage{british}
means `to get in touch', and here another homonym of the verb
\selectlanguage{russian}
\textit{выйти}
\selectlanguage{british}
is supposed to be):
\begin{figure*}[ht]
  \centering
  \includegraphics[width=\textwidth]{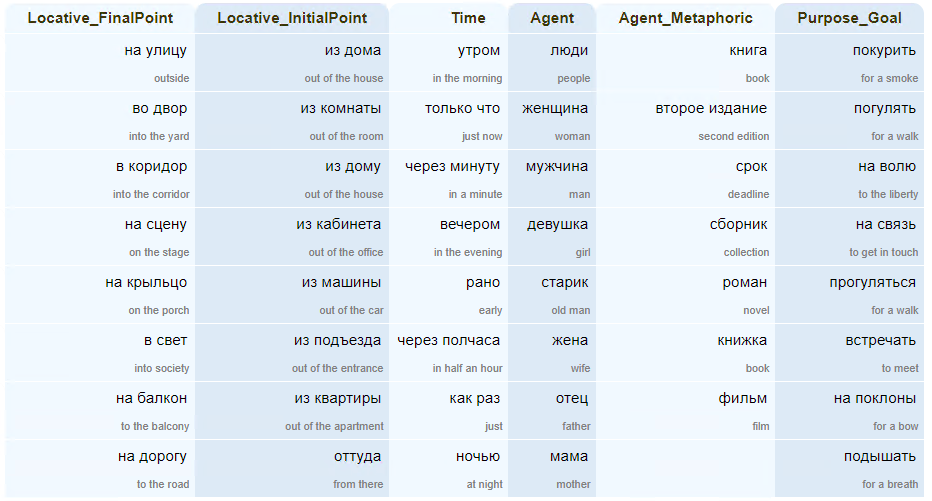}
  \caption{the semantic sketch for the verb 
  \selectlanguage{russian}
  <<выходить:идти:TO\_WALK>> 
\selectlanguage{british}
  `to go out'). Here the elements of the sketch are given with their rough translations.}
  \label{fig:walk}
\end{figure*}

The main reasons for the mistakes in the sketches are the incorrect influence of the statistics, certain inaccuracies of the semantic models in the parser, and the impossibility of distinguishing between the homonyms due to the closeness of their meanings or lack of distinguishing context in the sentences.

\subsubsection*{Contexts}
Every meaning from the chosen set is illustrated with contexts. A context is a sentence with one target predicate highlighted. No additional mark-up is presented.  Each meaning corresponds to several dozens of contexts with the target words having this meaning. The contexts were collected from news, fiction, publicistic texts,  
 being close by genre to those presented in Magazine Hall. It is important that the contexts do not overlap with the corpus which the sketches were built on. The excerpt from the contexts is given in Table \ref{context}.

\begin{table}
\centering
\begin{tabular}{l|l}
\textit{ID}      &  dev.sent.rus.116  \\
\textit{target}  & \selectlanguage{russian}наполнились 
\selectlanguage{british} \\
\textit{start}   &  46 \\
\textit{end}     &  57 \\
\textit{context} &  \selectlanguage{russian} Когда доктор вошел, она вспыхнула, и глаза ее наполнились слезами  
\selectlanguage{british}

\end{tabular}
\caption{The example of the context. The position of the target word 
\selectlanguage{russian}
\textit{наполнились} 
\selectlanguage{british}
`filled' in the context `When the doctor came in, she flushed, and her eyes filled with tears' is defined by the offsets. }
\label{context}
\end{table}

\subsubsection*{Datasets}
The task was meant to be solved in a few-shot or unsupervised learning manner. During the Shared Task, we provided the participants with two sets of data. In the first phase, the Trial data was published. It comprises three parts: a set of sketches, a set of contexts, and mapping between these two sets. The participants could use the data to get familiar with the formats, to test their hypotheses and to fine-tune their systems. During the second phase, we provided the participants with the main set of the sketches and corresponding contexts, which will be referred to as Dev data. 

In contrast to trial data, where the mapping had been given, for Dev data the participants were asked to find the relations between the sketches and the contexts themselves.

For the third phase, we manually selected 100 sketches and evaluated the corresponding contexts. This data formed the gold standard set for the task, which we will refer to as Manual Dev data.
Table \ref{splits} shows the size of the obtained datasets.

\begin{table}
\centering
\begin{tabular}{l|lll}
\textbf{split}      & \textbf{number of sketches} & \textbf{number of contexts} &   \\ 
\cline{1-3}
\textit{Trial}      & 20                 & 2000               &   \\ 

\textit{Dev}        & 895                & 44750              &   \\
\textit{Manual Dev} & 100                & 4347               &  
\end{tabular}
\caption{The size of the SemSketches datasets, \textit{Manual Dev} data forms a  subset  of \textit{Dev} data}
\label{splits}
\end{table}

During the second phase, the participants were able to commit their answers to the CodaLab\footnote{\url{https://competitions.codalab.org/competitions/29992}}  to know the results on the Dev data and to choose the best decision. During the third phase, the performance of the best variants was finally evaluated on the Manual Dev data.

After the announcement of the results, we published the answers (the mapping between the sketches and the contexts) on the SemSketches github. 

\subsection{Evaluation metric}
The submitted systems were evaluated using the \textbf{accuracy} metric. 
For the Shared Task, accuracy was calculated as the number of matched pairs between the participants' answers and test markup divided by the total number of contexts.

The evaluation script is publicly available on the SemSketches github.

\subsection{Baseline}
The participants were provided with a weak baseline solution.
The solution was based on the masked language modeling (MLM) mechanism of the RuBERT \cite{rubert} model.

For a given context $\textit{cont}$, $\textit{sketch}$ was chosen according to the computed sketch scores based on MLM candidates.
MLM candidates ($\textit{MLM}_\textit{cont}^\textit{N}$) were calculated as follows:
\begin{enumerate}
    
    \item syntactic analysis using the UDPipe (\cite{udpipe:2017}) has been performed to find the direct dependents of the target predicate;
    \item for each of the direct dependents, top-$\textit{N}$ mask replacements $\textit{Rep}_\textit{dep}^\textit{N}$ were stored;
    \item stored replacements were intersected, i.e. 
    $\textit{MLM}_\textit{cont}^\textit{N} = 
     \bigcap \{\textit{Rep}_{dep}^{N} \ \forall \textit{dep} \in \textit{cont}\}$;
    \item sketch $\textit{Score}$ was computed as the number of tokens present in the intersection of the sketch representation and the stored MLM candidates.
\[
    \textit{Score}\left(\textit{sketch},\textit{cont}\right) = \left|\textit{MLM}_\textit{cont}^\textit{1000} \cap \textit{Tokens}_\textit{sketch}\right|
\]
\end{enumerate}

The intersection was performed over lemmas thus treating 
\selectlanguage{russian}
\textit{на заре} 
and \textit{заря} 
\selectlanguage{british}
as intersecting entries.

The weak baseline system has shown 0,0094 accuracy on the Dev data set thus overperforming the random baseline.

\subsection{Submitted systems}
Three teams participated in the Shared Task: \texttt{paleksandrova}, \texttt{good501}, \texttt{smpl}. All teams suggested the solutions based on different approaches, and each solution managed to overcome the baseline. However, the final scores of each team turned out to be rather modest. To compare the results achieved, see Table \ref{tab:results} where the score of each team and the baseline score are presented. 

\begin{table}[ht!]
    \centering
	\begin{center}
	{\begin{tabular}{|r|l|l|}
		\toprule 
		\textbf{Team} & \textbf{Dev Score} & \textbf{Manual Dev Score}\\
		\cline{1-3}
		\texttt{paleksandrova}& $0.309$  & $0.277$ \\
		\texttt{good501} & $0.104$ & $0.127$ \\
		\texttt{smpl}& $0.182$  & $0.121$ \\
		\texttt{baseline} & $0.0094$ & $0.0035$ \\
		\bottomrule
	\end{tabular}}
	\caption{SemSketches Shared Task: the results of the submitted and baseline systems.}
	\label{tab:results}
    \end{center}
    
\end{table}

Let us now shortly characterize each decision and analyze what core problems they faced.

\paragraph{The team \texttt{smpl}} used the brute-force approach. LM score has been used to rank sketches and choose the best one for each context. 
To estimate how well the predicate $\textit{pred}$ fits into the given sketch $\textit{sketch}$, the $\textit{LM score}$ was used. 
$\textit{LM score}$ is
the average probability of $pred$ to replace \textit{[MASK]} token in template sentence `\textit{[MASK] cell}'. Template sentences were generated for each $\textit{cell}$ present in $\textit{sketch}$.


\paragraph{The team \texttt{Good501}} used the approach based on the sentences' similarity objective, which is a popular objective when training language models.
Target predicate was highlighted in the sentence using special tags. 
Sketch tables were flattened into pseudo-sentences.
For the given sentence, the most similar sketch was chosen by using the Sentence-BERT\citep{reimers-gurevych-2019-sentence} siamese similarity mechanism.


\paragraph{The team \texttt{paleksandrova}} \cite{winning_paper} used the MLM approach, which consisted of first restoring the covered predicate for each of the given sketches, and then picking the relevant sketch for the target sentence. 

The covered predicates were restored by generating templates (e.g. 
\selectlanguage{russian}
<<\textit{[MASK] в школу}>>
\selectlanguage{british}
--- `[MASK] to school')  using the sketch content cells. 
The most frequent predicate of all the MLM hypotheses for the sketch's templates was treated as the re-covered predicate. 
The first sketch whose predicate matched the sentence predicate was used as the system answer. When no sketch was found by exact matching, the sketch whose restored predicate was word2vec-closest \citep{Mikolov13} to the sentence predicate was used as the answer.

%

\subsection{The analysis of the submitted systems}

During the Shared Task, we formulated the experimental problem leaving enough room for different approaches. Although the performance of the submitted systems may be improved significantly, the proposed ideas were encouragingly diverse and thought provoking. 
The common feature of all three  systems is using the pretrained Language Models.

\paragraph{The team \texttt{501good}} which adopted the approach from  Sentence Transformers introduced the only system that included training. The model was trained on the \textit{Trial} data (20 sketches).

The systems of \texttt{smpl} and \texttt{paleksandrova} defined their unsupervised strategies for mapping the sketches and the contexts.
While the \texttt{smpl} team estimated how well each target predicate fits to each sketch using the score from the Masked Language Model, the \texttt{paleksandrova} team suggested an original approach imitating the way humans guess the core of the anonymous sketch. 

It is worth mentioning that the approaches of \texttt{paleksandrova} and \texttt{smpl}  by design cannot disambiguate the polysemous predicate, as they take  only the target verb into account but not its context.

\paragraph{The team \texttt{smpl}} approach could be thought of as scoring how well the sketch could account as the sentence predicate core. 
LM is trained on sentence-level objective, therefore, the successful application of the similarity approach demands more sophisticated preprocessing of the input sequence, for example, taking the predicate contexts into account. Such modification could improve the results.

\paragraph{The team \texttt{paleksandrova}} approach seems to be the most promising one. But the accuracy turned out to be rather low for the following reason. The sketch accumulates several verb forms, namely, it includes all tense, aspect and voice forms. For instance, the verbs 
\selectlanguage{russian}
\textit{строить}
\selectlanguage{british}
`build' \textit{Imperfective,  NonReflexive}, 
\selectlanguage{russian}
\textit{построить}
\selectlanguage{british}
`build' \textit{Perfective, NonReflexive},
\selectlanguage{russian}
\textit{строиться}
\selectlanguage{british}
`build' \textit{Imperfective, Reflexive},
\selectlanguage{russian}
\textit{построиться}
\selectlanguage{british}
`build' \textit{Perfective, Reflexive}
refer to one sketch.
As far as \texttt{paleksandrova} approach is concerned, the team regarded such verbs as different candidates for a sketch. At the same time, they chose only one top candidate for each sketch. Therefore, only one grammatical form of the necessary set could be referred to the right sketch.

\section{Discussion}

In the current paper, we demonstrated the pilot corpus of the semantic sketches, gave a brief analysis of the problems we faced during the corpus creation, and described the results of the SemSketches Shared Task aimed at applying the machine processing tools to the corpus.

The sketches are based on the parser with full semantic mark-up, which defines their value and uniqueness: first, the sketches allow one to analyze not only the actant dependencies, but a full semantic model of a word; second, they differentiate between the various meanings of the verbs.

As far as the opportunities for theoretical investigations are concerned, the sketches can help in dealing with all problems bound with the semantic compatibility of words. Especially, the SRL and the WSD problems must be mentioned here. 

As noted above, most researchers focus mainly only on the actant roles, while other dependencies do not usually get much attention. The semantic sketches suggest interesting data in this respect. The sketches include most frequent collocations, that is, the most natural, most typical contexts of a word. Among the dependencies the sketches include, modifiers and adjuncts are quite frequent. For some verbs, they seem to be even more specific than the actants and give more help in identifying the predicate.

For instance, the Locative is a typical circumstantial adjunct, but it is an obligatory slot for the verbs with the position meaning such as 
\selectlanguage{russian}
\textit{быть}
\selectlanguage{british}
`be', 
\selectlanguage{russian}
\textit{находиться}
\selectlanguage{british}
`be situated'. The Locative slot helps to differentiate between the `be' with the position meaning and other be-homonyms, while the semantic role corresponding syntactically to the Subject of `be' does not really contribute to differentiating between the be-homonyms.

It seems that the meaning of the adjuncts and the modifiers is sometimes underevaluated, therefore, an interesting task is to evaluate the correlation between the actant and circumstantial dependencies in the sketches.

As for the applied tasks, one of the promising directions in using the semantic sketches is their implementation for probing tasks for the pretrained language models. The interpretation of the linguistic knowledge encoded in the pretrained models has attracted much attention recently (\citep{conneau-etal-2018-cram}, \citep{Vuli2020ProbingPL}, \citep{ravichander2021probing}). We believe that the semantic sketches can serve as a basis for both  probing tasks and linguistically-motivated fine-tune tasks for such models.

To summarize, the  ideas from the proposed approaches can be used to embed effectively semantic sketches, making them not only a tool for manual lexicographical work but a semantic representation valid for automatic methods of Natural Language Processing.

\section{Further plans}

Our next plan is to add the sketches into the GICR, which brings two problems to consider.

The first one deals with the errors evaluation: in the current work, we did not check all the sketches in the pilot corpus manually --- only the manual Dev data. Therefore, we did not evaluate the total number of the mistakes in the whole corpus. This task is still to be done, including work on both, that is, sketches that seem to be unsuitable (checking the manual Dev data shows that such cases are rare) and sketches containing single mistakes in either the semantic dependencies or their fillers.

The second question is about the processing tools the sketches should be provided with. The SemSketches Shared Task demonstrated that machine tools can be successfully applied to the sketches processing (in spite of the fact that the precision of the solutions suggested by the applicants was not really high). 
What the tools should look like, depends significantly on the tasks the sketches will be used to solve. 

At the same time, we have recently started work on the English sketches, so our further plans include adding other languages to the sketch model, starting with the English sketches.

\bibliography{dialogue.bib}
\bibliographystyle{ugost2008ls}

\end{document}